\documentclass[letterpaper, 10 pt, conference]{ieeeconf}  

\IEEEoverridecommandlockouts                              

\usepackage{afterpage}
\usepackage{algorithm}
\usepackage[]{algpseudocode}

\usepackage{stix}
\usepackage{amsmath}

\DeclareSymbolFont{cmsymbols}{OMS}{cmsy}{m}{n}
\SetSymbolFont{cmsymbols}{bold}{OMS}{cmsy}{b}{n}
\DeclareSymbolFontAlphabet{\mathcal}{cmsymbols}

\usepackage{arydshln}
\usepackage[english]{babel}
\usepackage{bm}
\usepackage{caption}
\usepackage[T1]{fontenc}
\usepackage[]{graphicx}
\usepackage{hyperref}
\usepackage[utf8]{inputenc}
\usepackage{marginnote}
\usepackage{multirow}
\usepackage{multicol}
\usepackage{xcolor}
\usepackage{soul}
\usepackage{subfig}
\usepackage{tikz}
\usepackage{url}
\usepackage[backend=biber,style=ieee,sorting=none]{biblatex}
\usepackage{fancyhdr}

\newcommand{\trsp}{{^{\top}}}

\newcommand*\circled[1]{\tikz[baseline=(char.base)]{
            \node[shape=circle,draw,inner sep=2pt] (char) {#1};}}

\usepackage[nolist]{acronym}
\newacro{smc}[SMC]{sensorimotor contingency}
\newacro{dfc}[DFC]{dynamic functional connectivity}
\newacro{fc}[FC]{functional connectivity}
\newacro{nnmf}[NNMF]{non-negative matrix factorization}
\newacro{imi}[IMI]{instantaneous mutual information}
\newacro{irm}[IRM]{infinite relational model}
\newacro{mi}[MI]{mutual information}

\newacro{}[]{}
\newacro{}[]{}
\newacro{}[]{}

\title{\LARGE \bf
Unsupervised Discovery of Behavioral Primitives from Sensorimotor Dynamic Functional Connectivity
}

\author{Fernando D\'iaz Ledezma$^{1,\dagger}$, Valentin Marcel$^{2,\dagger}$, and Matej Hoffmann$^{2}$
\thanks{V.M. and M.H. were supported by Czech Science Foundation (GA CR), projects 20-24186X and 25-18113S.  F.D.L. acknowledges the funding by the Deutsche Forschungsgemeinschaft through the Gottfried Wilhelm Leibniz Programme (awarded to Sami Haddadin; grant no. HA7372/3-1). The work initiated during the stay of F.D.L. at CTU, supported by the OP VVV MEYS funded project CZ.02.1.01/0.0/0.0/16\_019/0000765.}
\thanks{$^{1}$Fernando D\'iaz Ledezma is with the Munich Institute of Robotics and Machine Intelligence, Technical University of Munich, Munich, Germany {\tt\small fernando.diaz@tum.de}}%
\thanks{$^{2}$Valentin Marcel and Matej Hoffmann are with the Faculty of Electrical Engineering, Czech Technical University in Prague, Prague, Czech Republic
        {\tt\small \{valentin.marcel,matej.hoffmann\}@fel.cvut.cz}}%
\thanks{$^\dagger$ The authors contributed equally to this work.}
}

\begin{document}
	
\maketitle
\thispagestyle{fancy}  

\begin{abstract}
The movements of both animals and robots give rise to streams of high-dimensional motor and sensory information. Imagine the brain of a newborn or the controller of a baby humanoid robot trying to make sense of unprocessed sensorimotor time series. Here, we present a framework for studying the dynamic functional connectivity between the multimodal sensory signals of a robotic agent to uncover an underlying structure. Using instantaneous mutual information, we capture the time-varying \ac{fc} between proprioceptive, tactile, and visual signals, revealing the sensorimotor relationships. Using an infinite relational model, we identified sensorimotor modules and their evolving connectivity. To further interpret these dynamic interactions, we employed non-negative matrix factorization, which decomposed the connectivity patterns into additive factors and their corresponding temporal coefficients. These factors can be considered the agent’s motion primitives or movement synergies that the agent can use to make sense of its sensorimotor space and later for behavior selection. In the future, the method can be deployed in robot learning as well as in the analysis of human movement trajectories or brain signals.
\end{abstract}
\section{Introduction}\label{sec:intro}
Structural connectivity describes the physical links between different components within a network. In neuroscience, for example, this usually refers to the anatomical pathways that connect various regions of the brain \cite{Park2013StructuralAF}. In contrast, \acf{fc} captures statistical dependencies or temporal correlations between the activities of different elements of the network. As described in \cite{Friston2011FunctionalAE}, \ac{fc} is an information-theoretic measure that quantifies these dependencies by analyzing the probability distributions of observed signals. Understanding the structural and functional connectivity of the brain has significantly advanced our knowledge of its organization and information processing capabilities. A similar approach can be applied to studying the sensorimotor signals of an embodied agent, offering insight into how the agent processes information and adapts its behavior. Although in the absence of body knowledge, analyzing the structural connectivity of these signals may not always be feasible, investigating their \ac{fc} provides valuable insight into how the agent understands its body structure and the emergence of behavior through information acquisition. 

The \acfi{smc} \emph{theory} \cite{oregan2001sensorimotor} provides an action-oriented account of cognition and consciousness, emphasizing the lawful relationships between actions and their sensory consequences. Picking up these relationships from high-dimensional streams of sensorimotor information is challenging. Information-theoretic measures, such as \ac{mi}, can be useful here, as they can find relationships between different variables in time. To gain maximum insight into what can be learned about the sensorimotor space using these methods, here we use an artificial agent consisting of a fixed torso and two flailing arms, inducing proprioceptive, tactile, and visual feedback. The situation is inspired by a newborn moving in a rather uncoordinated fashion and generating rich cross-modal activations, such as during self-touch or self-observation. We present a hierarchical framework that allows such an agent to discover its behavior primitives. The framework can be applied to robotics---where an agent could use these primitives for action selection---as well as to understand sensorimotor development of infants.

\begin{figure*}[h!]
	\centering
	\includegraphics[width=0.95\textwidth]{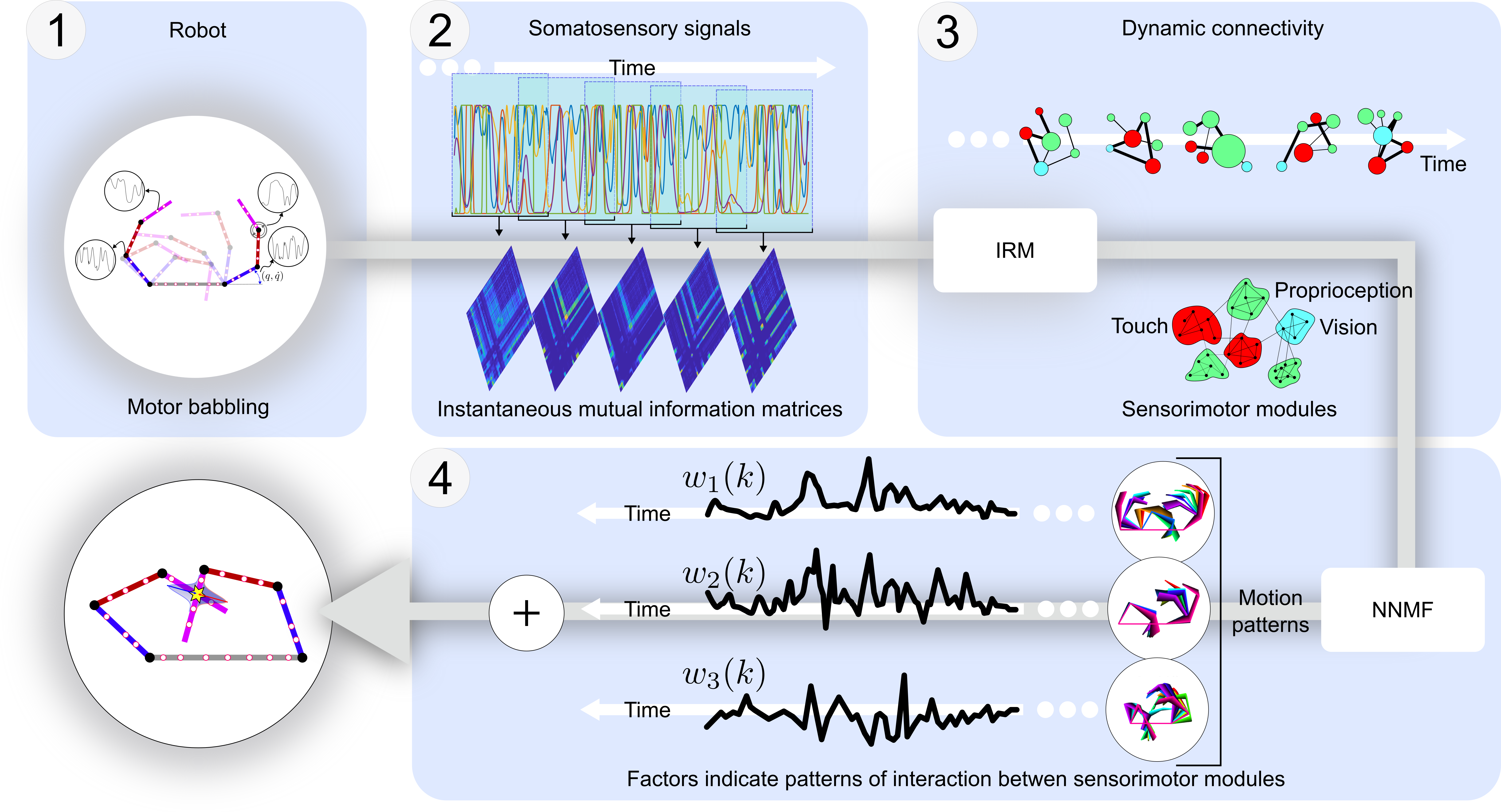}
	\caption{\textbf{From sensorimotor \acl{dfc} to behavior primitives.}~Motor babbling activates the agent's sensorimotor system. As it moves, \ac{imi} tracks the changing sensorimotor \ac{fc}. The \ac{irm} clusters pairwise \ac{imi} to reveal dynamic relations, which \ac{nnmf} then breaks down into behavior patterns.}    
	\label{fig:framework_overview}
\end{figure*}

\subsection{Related works}
Research suggests that \acp{smc} play a fundamental role in the acquisition of body knowledge, generalization, and goal-directed behavior \cite{Jacquey2019Sensorimotorcontingenciesas}. As such, they can be viewed as a form of sensorimotor representation, a framework that enables an embodied agent to learn, adapt, and interact with its environment. Despite extensive research on sensorimotor representations \cite{Nguyen2021Sensorimotorrepresentationlearning}, the precise relationship between sensorimotor regularities, body knowledge, and behavior remains an open question.

Several studies have used information-theoretic metrics to examine these relationships in sensorimotor systems. 
For example, \ac{mi} was used by D\'iaz et al. \cite{DiazLedezma2023MachineLS} to discover the mechanical topology of a robot from its proprioceptive signals. Touch has been identified as crucial for understanding \acp{smc} among different sensory modalities. Gama et al. \cite{Gama2021Goaldirectedtactile} demonstrated how intrinsic motivation and goal-babbling can facilitate self-touch learning in a simulated humanoid robot with artificial tactile skin. Similarly, Roncone et al. \cite{Roncone2014Automatickinematicchain} showed that self-touch could be used for kinematic calibration, allowing a robot to close its kinematic chain autonomously by touching its own body. Marcel et al. \cite{Marcel2022Learningreachown} further explored self-touch representation using a denoising framework with a multimodal variational autoencoder, enabling a robot to reconstruct its self-reaching configurations internally.

Another relevant area of study is \ac{dfc}, which explores how the statistical properties of sensorimotor signals evolve over time. Originally applied in neuroscience, \ac{dfc} has been used to detect states of reduced \acl{fc} during the onset of epilepsy \cite{Christiaen2020Dynamicfunctionalconnectivity} and to identify abnormal connectivity patterns in brains affected by disease \cite{Zhou2020Earlychildhooddevelopmental}. More recently, \ac{dfc} has been used to investigate how sensorimotor connectivity evolves in infants as they develop \cite{Kanazawa2023Openendedmovements}.

In robotics, \ac{fc} is expected to change based on the robot’s motion policy or the task it performs. Capturing and analyzing these evolving patterns using \ac{dfc} could provide a deeper understanding of how the sensory and motor systems of an agent interact over time.


\subsection{Overview}
This work introduces an analytical framework, illustrated in Fig.~\ref{fig:framework_overview}, to investigate the formation of relationships between the sensorimotor signals of a simulated embodied agent. Specifically, the framework examines the time-varying \acl{fc} between the agent's proprioceptive, tactile, and visual inputs. During an initial exploratory phase using motor babbling, the agent gathers information about its evolving sensorimotor relationships. The framework models these changing relationships as time-varying graphs, serving as a proxy to understand the formation and evolution of \acp{smc}. To achieve this, we compute the instantaneous pairwise \ac{mi} between sensorimotor signals, capturing the \acl{dfc} that correlates the agent's information sharing state with its movement and potential interactions. A key step in the framework involves applying a Bayesian approach to uncover hidden structures within mutual-information-based connectivity. This analysis identifies clusters of highly related signals and their evolving interactions over time. Finally, the framework decomposes these dynamic relationships to extract meaningful patterns (i.e., subgraphs) in the graphs, which represent distinct information-sharing states. The insights derived from these states and their transitions reveal fundamental aspects of the body structure and behavior of the agent.
\begin{figure}[!t]
	\begin{center}
		\hspace*{\fill}
		\includegraphics[width=0.95\columnwidth]{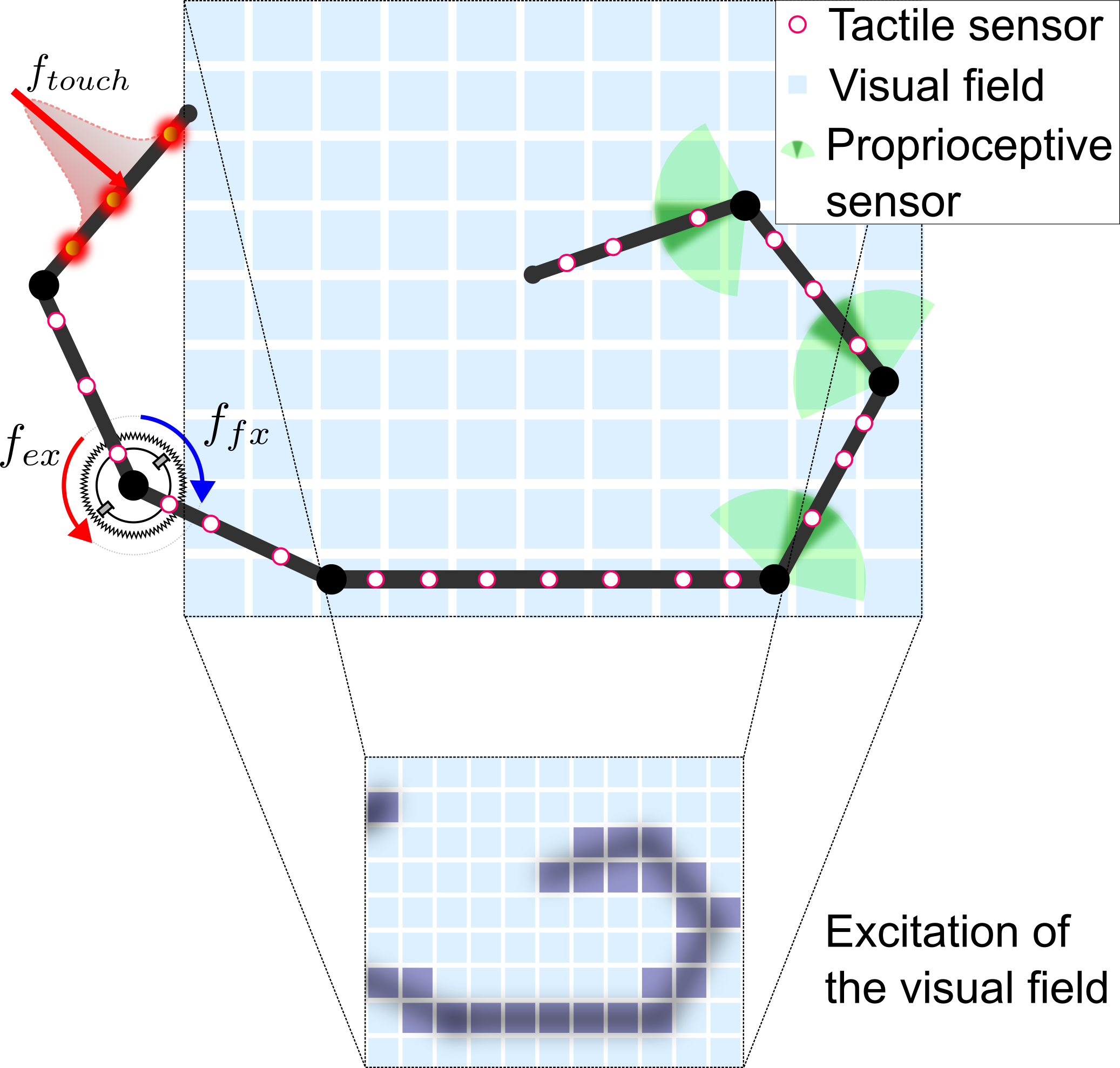}
		\hspace*{\fill}
	\end{center}
	\caption{\label{fig:extended_dual_arm_robot} \textbf{The embodied agent.}~The planar dual-arm robot uses antagonistic actuation at each joint. It receives tactile and proprioceptive input from the body and joints, while the visual field (blue area) detects body parts within its range.}
    \vspace{-20pt}
\end{figure}
\section{The embodied agent}\label{sec:the_embodied_agent}
\subsection{The planar dual arm model}
We started from the model described in \cite{Mannella2018Knowyourbody} and used in \cite{Marcel2022Learningreachown}, adding population coding to model sensory receptors. This model consists of a simple planar dual-arm system with six degrees of freedom, featuring three links per arm and a fixed torso, see Fig.~\ref{fig:extended_dual_arm_robot}. The robot is equipped with tactile sensors distributed throughout its body. 

In this work, we extend the model as follows. 
To instantiate the dynamics of the model, inertial properties are assigned to the robot's composing links. Its actuation mechanism is based on a biologically inspired model presented in \cite{Shim2012Chaoticexplorationlearning}, where the position $q$ and the velocity $\dot{q}$ of each joint are driven by antagonistic muscles (modeled as spring-damper systems). The pulling force these muscles exert is linearly controlled by the signal generated by a corresponding motor neuron $\sigma$. The joint torque,
\begin{equation}\label{eq:antagonistic_torque}
	\tau = \alpha \left(\sigma_\mathrm{fx} - \sigma_\mathrm{ex}\right)  + \beta \left(\sigma_\mathrm{fx} + \sigma_\mathrm{ex} + \gamma \right) q + \delta \dot{q},
\end{equation}
results from the difference between the activation signals for flexion $ \sigma_\mathrm{fx} $ and extension $\sigma_\mathrm{ex}$. These activation signals contribute to the pulling forces of flexion and extension, $ f_\mathrm{fx}$ and $f_\mathrm{ex} $, respectively. The remaining parameters of the model account for muscle force gain ($\alpha$), stiffness gain ($\beta$), tonic stiffness ($\gamma$) and damping coefficient ($\delta$).

\subsection{The sensory signals}
Tactile sensors are randomly distributed along the robot's body and are modeled using population coding \cite{Panzeri2010PopulationCoding}. Each sensor is represented by a Gaussian receptive field, whose mean is determined by the sensor's location. To incorporate touch strength, we modulate the activation of each receptive field based on contact force, adjusting the distance-dependent response accordingly.  

Similarly, the robot's proprioceptive measurements are also encoded using Gaussian receptive fields. In addition to somatosensation, the robot is equipped with visual inputs. Visual sensors consist of a fixed pixel receptive field with dimensions $(n_\text{x}, n_\text{y})$. When a limb segment intersects with a pixel in the visual field, the pixel value is set to one; that is, the pixels are sensitive to the positions of the agent's limbs. To account for the spatial overlap in some visual sensors~\cite{Marshall2015}, we apply a convolution operation using the 3-by-3 kernel \[
\small
K = \begin{bmatrix} 
0 & 0.25 & 0 \\ 
0.25 & 1 & .25 \\ 
0 & 0.25 & 0 
\end{bmatrix}
\] to the whole image followed by a saturation $\bm{v}=\texttt{min}(\bm{v},1)$ to keep the neural activations between $[0,1]$.

Ultimately, in our extended model, the visuosomatosensory signal vector consists of $N_\text{s}$ signals, including position-based proprioception ($\bm{p}$), tactile sensation modulated with touch strength ($\bm{r}$), and visual inputs ($\bm{v}$):  
\begin{equation}
	\bm{s} = \begin{bmatrix}
		\bm{p}\trsp & \bm{r}\trsp & \bm{v}\trsp
	\end{bmatrix}\trsp \in \mathbb{R}^{N_\text{s}}_{\geq}.
\end{equation}
In stage \circled{1} of our proposed analytical framework, the perceptual system of the embodied agent is stimulated to collect the signals $\bm{s}$ through a motor babbling strategy.

\section{The sensorimotor dynamic functional connectivity}

\subsection{\Acl{fc}}
\ac{fc} is a method for inferring network topology by characterizing the dependencies between observed signals based on their probability distributions \cite{Friston2011FunctionalAE}. Analyzing \ac{fc} helps uncover underlying structures that describe interactions between network entities. Building on the connection between embodiment and information structure, we hypothesize that an embodied agent’s structural properties and behavioral patterns 
can be revealed by studying the \ac{fc} among its sensorimotor signals $\bm{s}(t)$. 
~To quantify these relationships, we focus on \ac{mi}, an information-theoretic and model-free measure widely used for estimating linear and nonlinear dependencies between variables \cite{Steuer2002mutualinformationdetecting}.

The \ac{mi} between two random signals $I\left(X;Y\right)$ represents the amount by which a signal $ Y $ reduces the uncertainty about a signal $ X $. It is a symmetric measure of the information sharing between the two signals. By extension, the \ac{mi} matrix $\bm{\mathcal{I}} \in \mathbb{R}^{{N_\text{s}} \times {N_\text{s}}}$ can be constructed by computing the pairwise \ac{mi} between the $\left\lbrace s_i\right\rbrace^{N_\text{s}}_{i=1}$ sensorimotor signals. In practice, computing an entry $\left(\bm{\mathcal{I}}\right)_{i,j} = I(s_i;s_j)$~for a pair $\left({s}_i,{s}_j\right)$ involves centering their samples and using binning, kernel, or nearest-neighbor methods to compute their \ac{mi} \cite{WaltersWilliams2009Estimationmutualinformation}.

\subsection{\Acl{dfc}}
When analyzing \ac{fc}, it might be interesting to look not only at the aggregated effect of a complete dataset of recordings but also at the instantaneous changes that occur in the relationships. Indeed, the functional relationships between sensorimotor signals can change rapidly depending on the motion policy and the agent's interaction with the environment. To capture this time-varying, that is, dynamic, \acl{fc}, it is common to use a sliding time window with forward step $\Delta t$ from which the \ac{mi} is computed only for a small number of samples.

For a time window of length $T$, the \ac{mi} $I_t(s_x(k);s_y(k))$ between a distinct pair of signals $s_x(k)$ and $s_y(k)$ at time $k$ is calculated using the set of signal samples spanning the interval $\left[k-T,k\right]$. We refer to this quantity as the \acf{imi}. 

By extension, in stage $\circled{2}$ in Fig.~\ref{fig:framework_overview}, the \ac{mi} matrix $\bm{\mathcal{I}}(k)$ at time $t$ is constructed by calculating the \ac{imi} for all pairwise signals within the same time interval. The temporal evolution of this time-varying \ac{mi} matrix $\bm{\mathcal{I}}(k)$ captures the \ac{dfc} between the sensorimotor signals.

\subsection{Finding structure in the functional relationships}\label{sec:the_irm}
Analyzing the time-varying relationships in $\bm{\mathcal{I}}(k)$ at a local level may not be as informative as examining the global structure defined by interactions between groups of signals. To address this, we first identify clusters of closely related signals based on their \ac{mi} values. However, conventional clustering techniques typically require a predetermined number of clusters. To overcome this limitation, in stage \circled{3} of our framework, we employ a Bayesian approach---the \ac{irm}~\cite{Moerup2012}---to uncover hidden structures within the \ac{mi}-based connectivity.

This probabilistic framework uses relational data to group entities (e.g. signals) into $N_\text{c}$ clusters while simultaneously learning the relationships between them. In addition, it predicts new relationships based on the cluster assignments. The \ac{irm} utilizes a Chinese Restaurant Process or a Dirichlet Process, allowing the number of clusters to expand dynamically according to the data. Using probabilistic methods, the \ac{irm} estimates the likelihood of entities belonging to the same cluster and models intercluster relationships using probability distributions.

In our framework, \ac{irm} takes a thresholded and binarized time series of \ac{imi}-matrices---representing $N$ samples $\left\lbrace \bm{\mathcal{I}}(k)\right\rbrace^{N}_{k=1}$---and assigns to the i-th signal in $\bm{s}$ a cluster $z_i$. To identify these latent cluster relationships, \ac{irm} assumes that the probability of a relationship between two entities depends only on their cluster memberships. Specifically, if two entities belong to the clusters $z_i$ and $z_j$, the probability of connection is determined by a parameter $\theta_{z_i,z_j}$. These intercluster probabilities form a new binary relational matrix $\bm{Z}\in \mathbb{R}^{N_\text{c}\times N_\text{s}}$, which the model infers from the data. Furthermore, \ac{irm} produces a matrix array $\bm{H}\in \mathbb{R}^{N_\text{c}\times N_\text{c} \times N}_{\geq}$, which encodes the relationships between clusters, specifying the likelihood or strength of interactions between entities in different clusters.

\subsection{Analyzing recurring patterns}
The matrices $(\bm{Z}, \bm{H})$ in Sec.~\ref{sec:the_irm} capture the global changing likelihood of the relationships among the sensorimotor modules. In stage $\circled{4}$, the goal is to utilize the $N$ relationships in $\bm{H}$ to identify recurrent patterns in the \ac{fc} that correspond to distinct motor behaviors of the agent. Their significance is then determined by evaluating the strength and frequency of the patterns.

Common approaches for detecting repeating patterns in \ac{dfc} include the cosine similarity~\cite{Menon2019comparisonstaticdynamic}, k-means clustering~\cite{Li2017Hightransitionfrequencies}, and \ac{nnmf}~\cite{Fu2019Nonnegativematrixfactorization}. We selected the latter due to its demonstrated effectiveness in analyzing dynamic functional brain networks and its application in community detection~\cite{Luo2021Symmetricnonnegativematrix}.

To use \ac{nnmf}, a matrix $\bar{\bm{H}} = [\bar{\bm{h}}(1) \cdots \bar{\bm{h}}(k) \cdots \bar{\bm{h}}(N)] \in \mathbb{R}^{N_\text{r}\times N}$ is constructed by vectorizing and concatenating each of the matrices in $\bm{H}$ (i.e. a flattened version of the upper triangular part of $\bm{H}$); where $N_\text{r} = N_\text{c}(N_\text{c}-1)/2$. Since this matrix is strictly non-negative, it is amenable to be decomposed and analyzed using \ac{nnmf}. Then, the \ac{nnmf} algorithm can be used to factor the non-negative matrix $\bar{\bm{H}}$ into two parts: a matrix $\bm{F} \in \mathbb{R}^{N_\text{f}\times N_\text{r}}_{\geq}$ of the $N_\text{f}$ basis (or factors) and a matrix $\bm{W} \in \mathbb{R}^{N\times N_\text{f}}_{\geq}$ of their corresponding contributions or scores, such that
\begin{equation}
    \bar{\bm{H}}^{\top} \approx \bm{W} \bm{F}.
    \label{eq:nnmf}
\end{equation}

The \ac{nnmf} algorithm attempts to minimize the residual: 
\begin{equation}    
    D=\dfrac{1}{\sqrt{N_\text{r} N}}\|\bm{H}^{\top}-\bm{W} \bm{F}\|_F.
    \label{eq:residual}
\end{equation}

Reiterating, \ac{nnmf} decomposes the sensorimotor \ac{dfc} in $\bm{H}$ into: (1) a set of overlapping patterns $\bm{F}$---i.e., functional subnetworks---that evolve over space and time and (2) the corresponding coefficient time series $\bm{W}$ that indicate the contribution of each subnetwork at a given time.

\section{Results}
The model of the agent and the source code for our experiments are available
in a public repository\footnote{\url{https://github.com/ctu-vras/dfc-behavioral-primitives }}.
\subsection{Data collection}
\begin{figure*}[t!]
    \centering
    \includegraphics[width=0.95\textwidth]{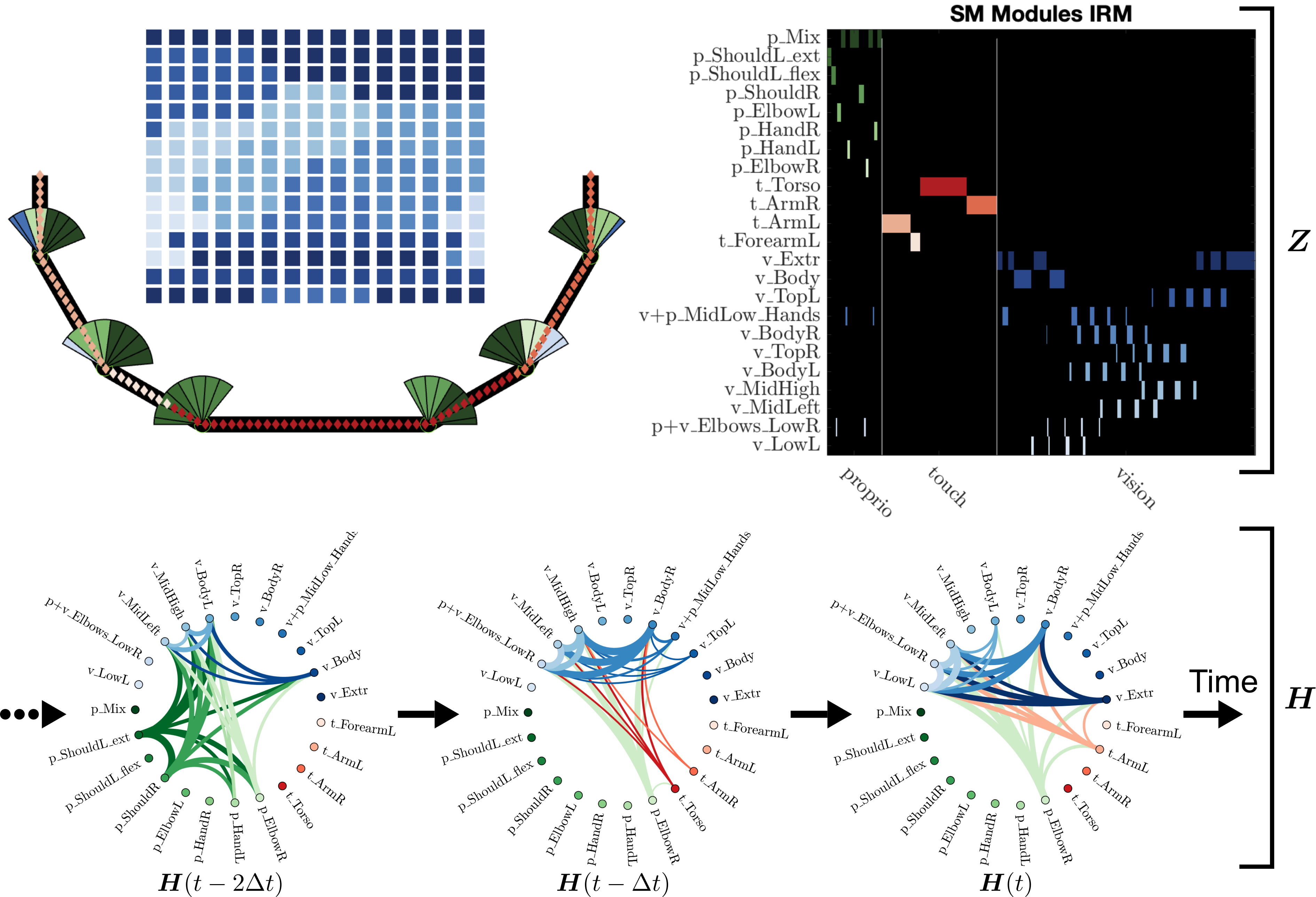}
    \vspace{5pt}
    \caption{\textbf{Extracted sensorimotor modules and their \ac{dfc}.}~\textsc{Right}: $N_\text{c}=23$ functional modules found by the \ac{irm}, each shown in a different color. Modalities (proprioceptive, tactile, visual) are clearly separated. \textsc{Left}: The agent's receptive fields colored by their module assignment. \textsc{Bottom}: Matrix $\bm{H}$ from the \ac{irm} shows \ac{dfc} between modules.}
    \label{fig:irm_modules}
    \vspace{-15pt}
\end{figure*}

The agent in Sec.~\ref{sec:the_embodied_agent} was simulated in MATLAB for a total time of 30 seconds with a sampling time of 1 ms (which corresponds to $N = 30,000$ data samples). Our experiment used a simple motor babbling strategy to stimulate the perceptual system and detect sensorimotor relationships by \ac{imi}. Each joint antagonistic pair received a periodic muscle activation command:
\begin{equation}\label{eq:motor_babbling_torques}
	\sigma(t) =  \text{tanh} \left( A_1 \text{sin}\left(\omega_0 t\right) + A_2 \text{sin}\left(2\omega_0 t\right) + A_3 \text{sin}\left(4\omega_0 t\right) \right);
\end{equation}	
with $A_i \sim \mathcal{U}(-1,1)$, $\omega_0 = 2\pi/T_0$, and $T_0=2$ seconds.

\subsection{Sensorimotor modules and \ac{dfc}}
To compute the \ac{imi}, we used sensor signals sampled at 100 Hz and a sliding window of $T = 0.1$ seconds, i.e., the previously seen 10 samples. This short memory was stored in a buffer. The selection of a relatively short time window is motivated by tactile events often occurring within a short timescale. In this work, we used a binning strategy to compute the matrix $\bm{\mathcal{I}}$ at each time. For the actual computation of the \ac{mi}, we used the open-source MATLAB package \emph{Mutual information computation} \cite{PengMutualInformationcomputation}. Consequently, to compute the clusters $\bm{Z}$ and the link densities $\bm{H}$, we used the \texttt{IRMUnipartite.m} function in the MATLAB-written package for Bayesian community detection available in \cite{Morup2025IRM}. 

Fig.~\ref{fig:irm_modules} depicts the output $\bm{Z}$ of the \ac{irm}. It found $N_\text{c} = 23$ functional modules corresponding to clusters of sensory signals with high information sharing (i.e., high \ac{imi}). We generally observe a clear separation of the sensory modalities into proprioception, touch, and vision (color-coded \textsc{green}, \textsc{red}, and \textsc{blue}, correspondingly). A few modules clustered visual and proprioceptive information together. Such modules originate from situations when the visual location of the hands is solely related to specific joint angle activations. 

It is worth noting the uneven number of elements across the modules. For example, the number of elements in the proprioceptive clusters in Fig.~\ref{fig:irm_modules} is smaller than in visual or tactile modules. This indicates that the signals in the proprioceptive modules carry a higher information-sharing level than modules that group a larger number of signals from other sensory modalities.

Once the functional modules have been extracted, we can compute the link densities $\bm{H}$, that is, the probability of having high \ac{mi} between the nodes in each module. For each 100 ms time window, we obtain a graph whose nodes are the sensorimotor modules and edges are link densities, depicted on the bottom panel of Fig.~\ref{fig:irm_modules}.

\subsection{Factors of the \ac{nnmf}}
\begin{figure*}[h!]
    \centering
    \includegraphics[width=0.99\linewidth]{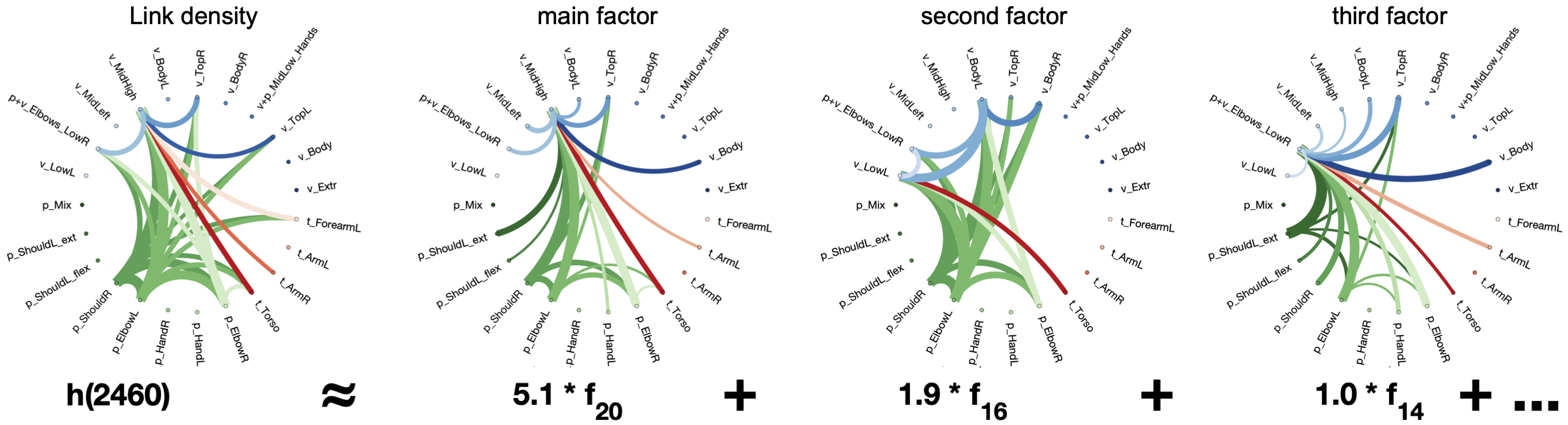}
    \caption{\textbf{Example of factor decomposition of the link density matrix based on \ac{nnmf}.} At sample 2,460, the agent's hands touch each other, as seen in Fig.~\ref{fig:factors_time}. The decomposition shows that factors $\bm{f}_{20}$, $\bm{f}_{16}$, and $\bm{f}_{14}$ contribute the most to this beavior.}
    \label{fig:decomposition}
\end{figure*}
\begin{figure*}[!t]
	\begin{center}
		\hspace*{\fill}
        \includegraphics[width=1\textwidth]{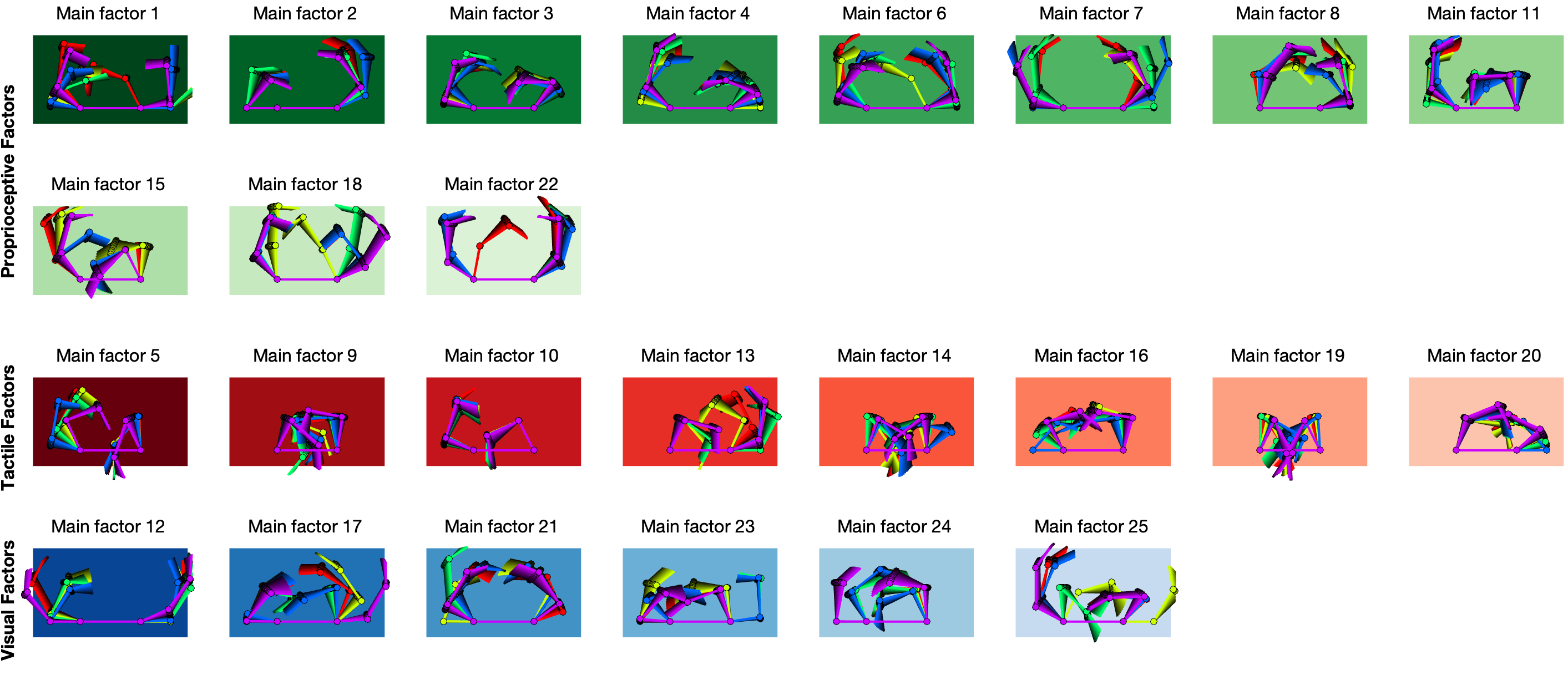}
		\hspace*{\fill}
	\end{center}
    \vspace{-25pt}
	\caption{\textbf{The factors and their associated behaviors.} Different events during exploration are linked to a given factor; for example, pure proprioception (no touch), contact with the left arm, the right arm, and both arms.}
    \label{fig:main_factors}
\end{figure*}
 
Factorizing the vectorized link densities matrix $\bar{\bm{h}}(k)$ using \ac{nnmf} leads to a set of basis factors  $\bm{F} = [ \bm{f}_1 \cdots \bm{f}_i \cdots \bm{f}_{N_\text{f}}]$ and their corresponding scores $\left\lbrace w_i(k)\right\rbrace^{N_\text{f}}_{i=1}$ that approximate the link densities as
\begin{equation}
    \bar{\bm{h}}(k)\approx \sum_{i=1}^{N_\text{f}} w_i(k) \bm{f}_i.
    \label{eq:decomposition}
\end{equation}
To choose the number of factors $N_\text{f}$, we follow the elbow method on the residual---Eq.~\eqref{eq:residual}---as in \cite{Phalen2020Nonnegativematrix}, and obtain a set of $N_\text{f} = 25$ nonnegative factors with a mean squared residual of $D=0.1658$.
An example of the decomposition is shown in Fig.~\ref{fig:decomposition}, where the link density connection is represented as a graph with edges thickness relative to the connection strength between the modules. The factors in $\mathbf{F}$ are ordered by their contribution to the reconstruction of the link densities matrix, which is computed as the total score energy $E_i =  \sum_{k=1}^N w_i^2(k)$, so that factors $\bm{f}_1$ and $\bm{f}_{25}$ have the highest and lowest contribution, respectively.

Each factor represents a specific type of connection between modules. They connect modules of touch to modules of proprioception, modules of vision to modules of touch, etc. It is possible to classify each factor based on their interacting modalities. For instance, the factors $\bm{f}_5$, $\bm{f}_9$, $\bm{f}_{10}$, $\bm{f}_{13}$, $\bm{f}_{14}$, $\bm{f}_{16}$, $\bm{f}_{19}$, $\bm{f}_{20}$ all include a connection to a tactile module. In contrast, other factors include specific visual connections, while others consist mainly of connections between proprioceptive modules. A key insight is that factors can be thought of as \emph{behavioral descriptors}. They decompose the agent's motion into synergies between sensory inputs that can be added together. Fig.~\ref{fig:main_factors} shows example motions and their associated leading factor (that with the highest score). 

\subsection{Factor scores as behavioral descriptors}
\begin{figure*}[h!]
    \centering
    \includegraphics[width=.99\textwidth]{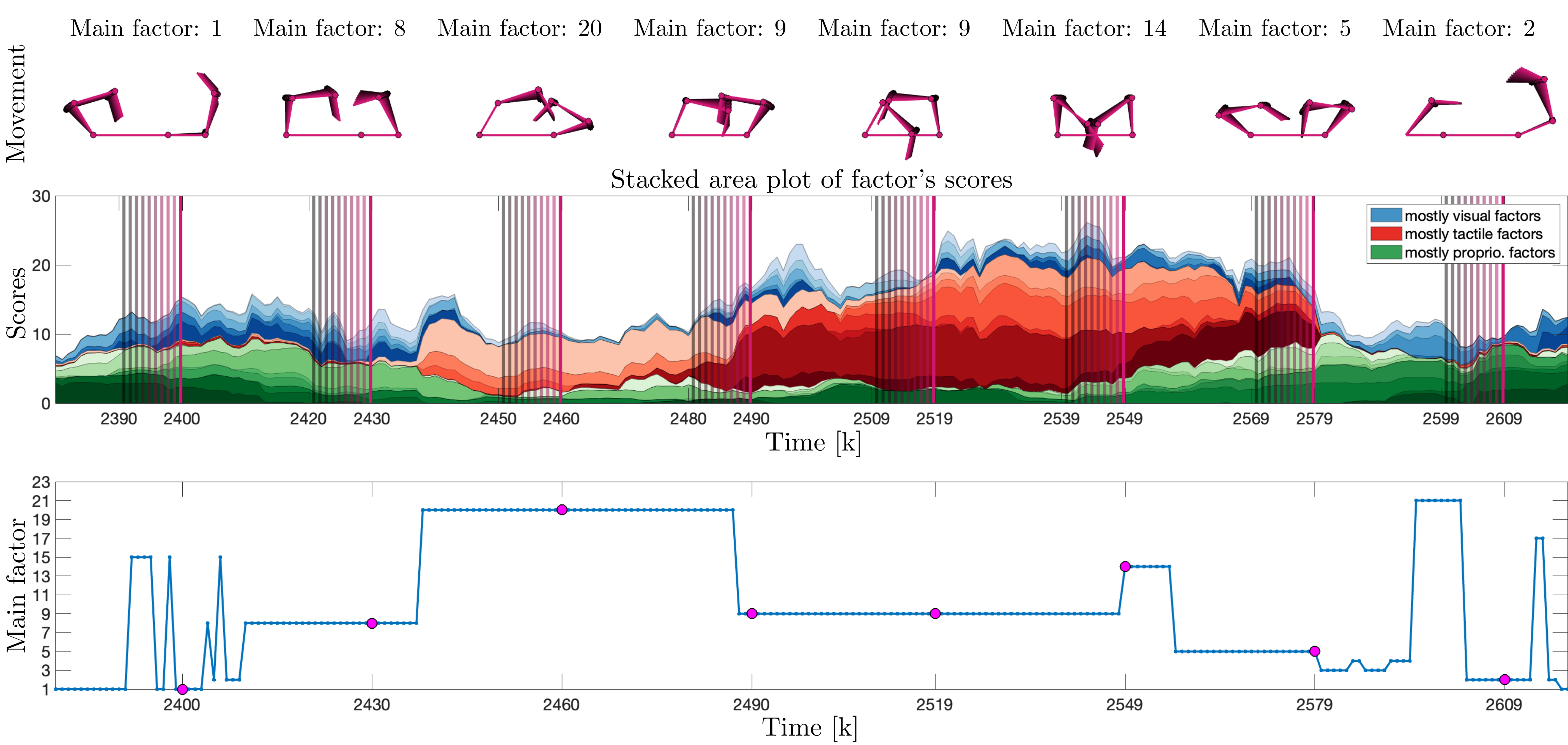}
    \caption{\textbf{Factors' scores, main factors, and agent's behavior.}~\textsc{Top}: Agent’s behavior at 8 time points, each spanning 10 samples—the window used for computing \ac{imi}---shown in pink from darker (earlier) to lighter (later). The sequence moves from arm waving to hands touching, hands on torso, one hand on torso, and back to no-touch. \textsc{Middle}: Stacked plot of factor scores. Factors with strong responses to visual, tactile, and proprioceptive modules are shown in blue, red, and green, respectively. \textsc{Bottom}: Time course of the main factor, i.e., the one with the highest overall contribution.}    
    \label{fig:factors_time}
\end{figure*}
The \acl{nnmf} has the property of extracting characteristics from the data such that each factor found represents a unique body synergy. In Fig.~\ref{fig:factors_time}, we observe the evolution of the scores and main factors during the agent's movement\footnote{A complementary video can be found in \url{https://youtu.be/Dr4y03FPcQc}.}. To show another property of the combination of \ac{irm} and \ac{nnmf} for the analysis, consider that while visual sensors are the most represented in raw neural signals with 225 neurons (compared to 48 proprioceptive neurons), the contribution of vision is quite reduced, as reflected in scores $\bm{W}$. Our proposed process normalizes the density of neural activation by their functional and informative content to describe the current behavior; as an example, such a property would provide a way to weigh sensory prediction errors in sensorimotor learning methods.

\section{Conclusion}
This work introduced a framework for analyzing the \acl{dfc} among an embodied agent's multimodal sensory signals to autonomously uncover the underlying structure in the sensorimotor streams it generates. 
From \acl{imi}, we captured the time-varying functional connections between proprioceptive, tactile, and visual signals, revealing fundamental sensorimotor relationships. 
Applying an \acl{irm}, we identified 23 sensorimotor modules and their evolving connectivity, represented by the link density matrix $\bm{H}$. 
To further interpret these dynamic interactions, we employed \acl{nnmf}, which decomposed the connectivity patterns into 25 additive factors (subgraphs, $\bm{f}_i$) and their corresponding temporal coefficients ($w_i(k)$).

Factors can be interpreted as a basis \ac{fc} graph, capturing distinct states of information sharing between sensorimotor clusters. These factors also encode fundamental movement patterns that represent the embodied interactions of the agent with its environment. They serve as proxies for \acp{smc} that emerge during motor babbling. At any given time, the observed interaction patterns of the agent can be viewed as a weighted combination of these \acp{smc}, governed by the entries of the coefficient matrix $\bm{W}$. By extracting structured representations of the sensorimotor dynamics of an embodied agent, our approach demonstrates the effectiveness of combining information-theoretic measures (\ac{mi}), Bayesian nonparametric models (\ac{irm}), and high-dimensional analysis (\ac{nnmf}) in studying \acp{smc}.

\section{Discussion}
How do our findings relate to \acp{smc}? This work provides a new approach to studying \acp{smc} by enabling an agent to autonomously organize its sensory space into functional modules. These modules, derived from the dynamic relationships between proprioceptive, tactile, and visual signals, form a structured representation---a ``patchwork'' of sensorimotor interactions. When combined, the connectivity patterns between modules describe the agent’s current state of sensorimotor interaction. This representation allows the agent to evaluate contingencies, such as whether certain sensory outcomes (e.g., touching the torso) depend solely on specific motor actions (e.g., shoulder flexion). Over time, the agent could infer rules such as: ``\emph{I cannot touch my torso without flexing my shoulder},'' or ``\emph{Waving my right elbow generates visual input on the right}.''

The patterns discovered through \ac{nnmf} can be interpreted as behavioral descriptors---sensorimotor synergies---that provide a compact, additive basis for encoding the agent’s experience. These factors serve as candidate motor primitives or latent behaviors that could be reused for planning or control.

While our framework focuses on the structure of sensory signals resulting from motor activity, it does not explicitly model motor or muscle commands. Including these in future work could yield more complete sensorimotor representations. Moreover, \ac{mi}, while effective for capturing non-linear dependencies, is symmetric and does not capture directionality. Alternative information-theoretic measures, such as transfer entropy or Granger causality, could provide insights into the causal flow of information.

The current setup uses random motor babbling to explore the sensorimotor space. Future work could incorporate goal-directed, curiosity-driven, or active exploration strategies to enrich the set of behaviors discovered. One promising direction is to apply the method to motion capture data from human infants, mapped onto biomechanical models \cite{Kanazawa2023Openendedmovements}, to analyze developmental sensorimotor structures in biologically plausible settings.

Finally, computational scalability remains a challenge. The dimensionality of the sensor space and the number of modules directly impact the cost of \ac{mi} estimation and clustering. Dimensionality reduction techniques could help mitigate this issue. Additionally, the current framework operates offline. Incorporating incremental methods such as online \ac{nnmf} \cite{Bucak2007} would enable real-time learning and adaptation. As the agent acquires new experience, novel factors may emerge while older ones fade, supporting continuous refinement of behavioral representations.

\printbibliography 
\end{document}